\documentclass[pmlr,anonymous=true]{jmlr}

\usepackage{jmlrutils}

\usepackage{longtable}

\usepackage{booktabs}
\usepackage[load-configurations=version-1]{siunitx} 

\usepackage{graphicx}

\usepackage{verbatim}
\usepackage{makecell}
\usepackage{float}
\usepackage{multirow}
\usepackage{authblk}

\theorembodyfont{\upshape}
\theoremheaderfont{\scshape}
\theorempostheader{:}
\theoremsep{\newline}

\begin{document}



\title[Query-Focused EHR Summarization to Aid Imaging Diagnosis]{Query-Focused EHR Summarization \\ to Aid Imaging Diagnosis}

\author[1]{\Name{Denis Jered McInerney} \Email{mcinerney.de@northeastern.edu}}
\author[2]{\Name{Borna Dabiri} \Email{bdabiri@bwh.harvard.edu}}
\author[2]{\Name{Anne-Sophie Touret} \Email{atouret@partners.org}}
\author[2]{\Name{Geoffrey Young} \Email{gsyoung@bwh.harvard.edu}}
\author[1]{\Name{Jan-Willem van de Meent} \Email{j.vandemeent@northeastern.edu}}
\author[1]{\Name{Byron C. Wallace} \Email{b.wallace@northeastern.edu}}
\affil[1]{Khoury College of Computer Sciences, Northeastern University, Boston, MA, United States}
\affil[2]{Brigham and Women's Hospital, Boston, MA, United States}
\editor{}
\maketitle

\begin{abstract}
  Electronic Health Records (EHRs)
  provide
  vital contextual information to 
  radiologists and other physicians when making a diagnosis.
  Unfortunately, because a given patient’s record may contain hundreds of notes and reports, identifying relevant information within these in the short time typically allotted to a case is very difficult.
  We propose and evaluate models that extract relevant text snippets from patient records
  to provide a rough case summary intended to aid physicians considering one or more diagnoses.
  This is hard because direct supervision (i.e., physician annotations of snippets relevant to specific diagnoses in medical records) is prohibitively expensive to collect at scale.
  We propose a \emph{distantly supervised} strategy in which we use groups of International Classification of Diseases (ICD) codes observed in `future' records as noisy proxies for `downstream' diagnoses.
  Using this we train a transformer-based neural model to perform extractive summarization conditioned on potential diagnoses. This model defines an attention mechanism that is
  conditioned on potential diagnoses (queries) provided by the diagnosing physician. 
  We train (via distant supervision) and evaluate variants of this model on EHR data from Brigham and Women's Hospital in Boston and MIMIC-III (the latter to facilitate reproducibility).
  Evaluations performed by radiologists demonstrate that these distantly supervised models yield better extractive summaries than do unsupervised approaches. Such models may aid diagnosis by identifying sentences in past patient reports that are clinically relevant to a potential diagnosis.
\end{abstract}

\section{Introduction}
\label{section:intro}

\begin{figure}
\centering 
\includegraphics[scale=.35]{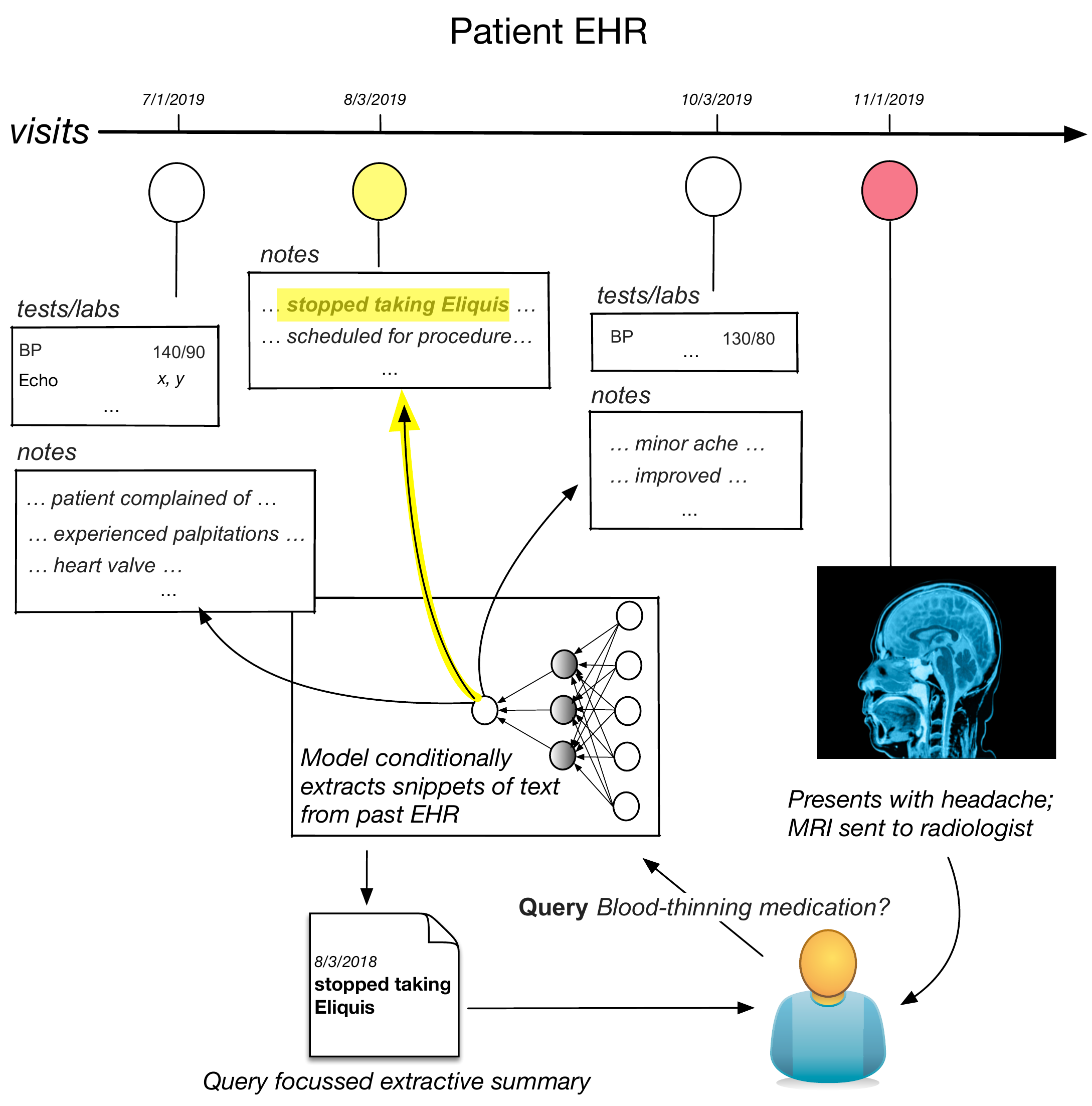}
\caption{We propose models for generating targeted, extractive summaries of the notes within patient EHR data to aid practitioners (here, radiologists) at point of care. We propose distant supervision schemes to train these models.}
\label{fig:overview}
\end{figure}

The transition to Electronic Health Records (EHR) has increased the clinical information about a patient available to providers by orders of magnitude. This clinical information takes the form of both structured fields (e.g.~lab tests) and unstructured (text) provider notes. The anticipated benefit of EHR is improved transparency and continuity of care between the various providers managing a particular patient. Unfortunately, in practice the amount of data in medical records can be overwhelming. 

A recent New Yorker article highlights the issue \citep{gawande2018doctors}: A primary care physician interviewed for this piece describes EHR as a ``massive monster of incomprehensibility''. Consequently: ``piecing together what’s important about the patient’s history is at times actually harder than when she had to leaf through a sheaf of paper records''. This makes it difficult --- and often impossible under existing time constraints --- for physicians to identify the information buried within a patients' EHR notes that might be critical to forming an accurate diagnosis. 

Here we consider the setting of radiologists interpreting medical imaging. Radiologists typically have fewer than 10 minutes to complete their interpretation, reporting, and communication of a case. Most of this time is necessarily spent inspecting the 1000s of images in a typical MRI or CT scan, analyzing abnormalities, formulating a diagnosis, producing a report, and communicating findings. This leaves practically no time to thoroughly consult patient history. 
Key information that might inform diagnosis is often only available in the notes within the EHR. However, the sheer volume of this unstructured information renders it nearly impossible for the radiologist to identify and capitalize on the relevant history. 
Radiologists thus currently interpret most medical imaging studies with little knowledge of the background information beyond the brief clinical indication listed in the request for a particular imaging study that is provided by the ordering physician.

This work aims to enable efficient use of patient history 
by presenting radiologists with relevant text that is automatically extracted from the EHR. 
Our focus on text complements the extensive body of work on image retrieval methods for diagnostic medical imaging \citep{kalpathy2015evaluating} (see Section \ref{section:related-work}), which aim to retrieve images similar to the one under consideration to inform diagnosis. Concretely, we develop neural natural language processing (NLP) models that extract compact summaries of textual diagnostic clinical information from patient EHR relevant to a given query. This interactive summarization system is intended to \emph{aid} in generation of a differential diagnosis; we envision the model presenting potential clues (text snippets from EHR notes) to the radiologist that might support different diagnostic possibilities. The radiologist will be able to \emph{query} the model to try and identify snippets of text (and numerical data) relevant to a particular potential diagnosis. Figure \ref{fig:overview} provides a schematic overview of the proposed approach.


Dearth of direct supervision poses a substantial challenge when training models to extract relevant text from clinical notes, particularly because we cannot hope to obtain annotation of relevant snippets with respect to every possible query. To train models in the absence of explicit supervision, we propose to use International Classification of Diseases (ICD) codes as a source of indirect supervision. By using ICD codes as noisy proxies for diagnostic labels, we are able to train models to extract snippets of text that are predictive of a future diagnosis.    

We propose and evaluate several models that can be conditioned on a given query (operationally, an ICD code or some representation of it). We introduce a novel Transformer-based architecture \citep{vaswani2017attention} that relies on a `pointer' mechanism (i.e., an attention distribution over inputs, \citeauthor{see2017get}, \citeyear{see2017get}) to produce extractive summaries. 
We train this model end-to-end to predict \emph{future} groups of ICD codes that correspond to diagnostic categories of interest. That is, we train the model to 
perform a soft scoring of relevant snippets in service of predicting downstream clinical outcomes codified in ICD codes, which are already available (i.e., require no manual supervision). Once trained, this model can be used as a mechanism to surface evidence relevant to arbitrary natural language queries.


Our radiology co-investigators evaluate variants of this model --- and simple completely unsupervised baselines based on cosine similarities between text encodings --- both retrospectively and prospectively on data from Brigham and Women's Hopsital, with the help of radiologist colleagues. On a small test set of manually collected annotations, variants of our weakly supervised model significantly outperform unsupervised baselines with respect to retrieval of relevant reference summaries, and in terms of clinicians' ratings of model outputs.


To summarize, the \textbf{main contributions} of this work are as follows. (i) We formalize the task of interactive, query-focussed extractive summarization of the notes within EHR to aid specialists (here, radiologists) performing diagnosis. (ii) We propose novel model variants that build on the state-of-the-art in neural summarization for performing this task. (iii) We propose several `distantly supervised' strategies to train these models, which differ in how they exploit the supervision signal, and in how they represent (derived) input queries. (iv) We perform realistic evaluations of extractive summarization models (including the proposed model and appropriate baselines) in which domain experts (radiologists) assess the utility of the summaries produced.
This contributes \textbf{generalizable knowledge} in that: (i) the proposed methods and distant supervision strategies for inducing targeted summaries of EHR for particular specialties may generalize to other specialities in medicine, and, (ii) the models we propose for query-focussed neural extractive summarization have more general applications for NLP in distantly supervised settings.

\section{Data and Cohort}




Our primary dataset is composed of EHR for patients who have received care at Brigham and Women's Hospital (BWH) in Boston. Because our senior radiologist co-investigator specialized in diagnostic neuroimaging, we retrieved all patients from the BWH database from 2004 to 2019 who had undergone magnetic resonance imaging of the brain at least once and had at least one visit to the emergency room. The latter constraint was designed to yield more patients undergoing MRI at a time when there was no definitive diagnosis.  This yielded 22,191 unique patients, of which 19,841 had at least one of the following: Discharge summaries, Operative reports, Pathology reports, Progress notes, Radiology reports, or Visit notes.  These notes comprise the text input into the model.


This data represents the actual target population for our work.  Unfortunately, we are unable to release the dataset publicly at present because of data privacy issues. To facilitate reproducibility and transparency, we therefore also perform experiments using the MIMIC-III dataset. The MIMIC-III patient population is not ideal for the proposed application since it consists of patients in the intensive care unit, but it is publicly accessible.

\subsection{Data Extraction}\label{section:data_extraction}

The raw data within EHRs comprise a series of tables of different reports and codes, as well as patient information. From this we assemble a list of different report types and diagnosis codes for each patient, using time stamps to sort these in temporal order. A radiologist on our team created a \textbf{hierarchy of diagnosis categories}\footnote{We will make this hierarchy available for public use with our code base and discuss it in more detail in the appendix.} in which the lowest-level categories (i.e.,~the leaves of the hierarchy) are associated with ICD codes that we treat as diagnosis codes. This hierarchy is used to relate codes that are similar and train general high-level diagnoses as well as more specific low-level diagnoses.  

For each patient, we characterize codes that appear at least twice as \emph{persistent}, constituting an ongoing condition. For each persistent code that can be mapped onto the hierarchy of codes manually designated to be of interest to our team radiologists, we search the EHR for a relevant radiology report within the year prior to the first occurrence of the persistent diagnosis. When such a relevant report is found an {\it instance} is created, defining a time-point $t$. The goal of the model is to summarize the EHR reports \emph{prior} to time $t$ by picking sentences that may serve as evidence for the onset of a persistent condition at time $t$. Each instance uniquely corresponds to a patient $p$ and time-point $t$.

More formally, we transform such instances into $(\boldsymbol{x}, \boldsymbol{q}, \boldsymbol{y})$ triples where $\boldsymbol{x}$ is a list of sentences, $\boldsymbol{q}$ a list of categories, and $\boldsymbol{y}$ is a list of labels indicating whether the constituent categories in $\boldsymbol{q}$ appear in the patient's `future' (with reference to $t$).\footnote{The model as outlined takes instances of the form $(\boldsymbol{x}, q, y)$ where $q$ and $y$ are an individual category and label, not lists of categories and labels. We simply average the loss over the list of categories and labels to obtain the final instance-wise loss.}
Each sentence is a list of tokens in the vocabulary $V$. The sentences in $\boldsymbol{x}$ are obtained by concatenating the reports before time point $t$ in a patient's record, splitting the concatenated records into sentences using the {\tt spaCy} english parser (\cite{spacy2}, version 2.2.3), and tokenizing these using the BERT Base Cased tokenizer from the {\tt HuggingFace} implementation (\cite{Wolf2019HuggingFacesTS}, version 2.4.1), or the {\tt sklearn} (\cite{sklearn}, version 0.22.1) TF-IDF tokenizer in the case of the TF-IDF similarity model. The categories in $\boldsymbol{q}$ are all categories for which we may want to predict a label given the records observed before time-point $t$. These include any persistent codes (leaf categories) that appear after time-point $t$ labeled as positive (1) in $\boldsymbol{y}$ and any codes that never appear in the patient's record labeled as negative.  Non-leaf categories are included labeled as positive if they have at least one positive descendant and labeled as negative if they have at least one negative descendant {\it and} no positive descendants.

We split patients into train, validation, and test groups with 13888, 2976, and 2977 patients respectively for the BWH dataset and 32302, 6922, and 6922, patients respectively for the MIMIC-III dataset. We create instances from these patient groups as detailed above, yielding 88856, 18726, and 18895 instances spanning 11004, 2338, and 2337 patients respectively for the BWH dataset, and 5158, 1123, and 1181 instances spanning 3736, 784, and 848 patients respectively for MIMIC-III. Due to memory constraints, we truncate the train set to 10,000 instances and the validation and test sets to 1,000 instances each.

\section{Methods}

\begin{figure}
\centering 
\includegraphics[scale=.5]{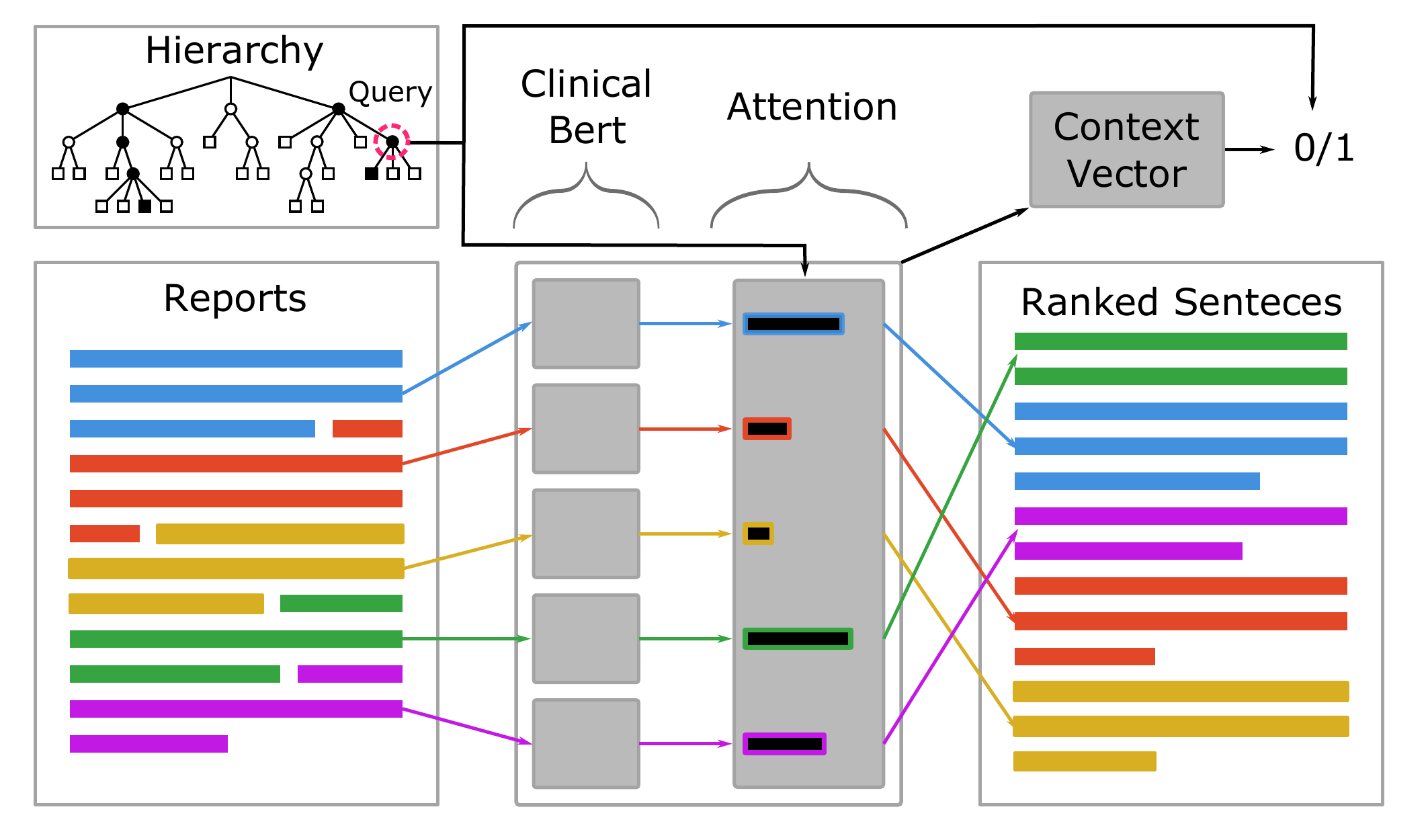}
\caption{Distant supervision model (Section \ref{section:distant-supervision}).  Queries are chosen at any depth of the hierarchy and can be either positive (filled in) or negative (empty) targets, which are predicted as 1 or 0 respectively.  The model also outputs a ranked list of sentences ordered based on the attention mechanism.}
\label{fig:model}
\end{figure}

\subsection{Models}
\label{section:models} 

Our aim is to design a model that takes as input (concatenated) free-text reports from a patients' medical record $\boldsymbol{x}$ and a \emph{query} of interest $q$; the model should then yield an extractive summary of $\boldsymbol{x}$ pertaining to $q$. Operationally, this entails inducing a soft distribution of `relevance scores' over the sentences comprising ${\boldsymbol{x}}$, which we will denote by $\boldsymbol{a}$. A summary can then be constructed by compiling the $k$ sentences corresponding to the highest scores in $\boldsymbol{a}$. Formally, we have:

\begin{equation}
    f_{\boldsymbol{\phi}}(\boldsymbol{x}, q) = \boldsymbol{a}
\end{equation}
where $\boldsymbol{\phi}$ are model parameters.

We would ideally train this model with ($\boldsymbol{x}$, $q$, $\boldsymbol{a}$) triplets provided by domain experts, but collecting a sufficient amount of this sort of explicit annotation would be prohibitively expensive. Therefore, we treat diagnosis ICD codes as proxy `queries' to induce distant supervision (see Section \ref{section:distant-supervision}).  
We consider three versions of categories as queries: category indicators (categorical queries); natural language descriptions to which these categories correspond, and; hierarchy embeddings, which concatenate natural language descriptions from nodes in the path leading to a category in the hierarchy (see Section \ref{section:query_embeddings}).
Below we describe the models we evaluate (including baselines) given this framing of the task.



\subsubsection{Unsupervised Baselines}
\label{section:baselines} 

We implement two unsupervised baselines for comparison to more complex (weakly) supervised models.
First, we define a \textbf{TF-IDF similarity} model that encodes all sentences in patient reports and the query (here, a natural language description associated with a given diagnosis category) into Term Frequency-Inverse Document Frequency (TF-IDF) Bag-of-Words (BoW) vectors, and uses the cosine similarity between these as a similarity score:
\begin{equation}
    \boldsymbol{a}_i = \mathrm{Cosine}(\mathrm{tfidf}(\boldsymbol{x}_i), \mathrm{tfidf}(\boldsymbol{d}_q))
\end{equation}
where $i$ indexes sentences,\footnote{These sentences are all of those corresponding to reports for a particular patient, but we elide the patient index here for clarity.} and $\boldsymbol{d}_q$ is the description that accompanies category $q$.\footnote{These are part of the ICD resource. Note that it does not make sense to attempt to use only the category indicator itself for unsupervised BoW approaches based on similarities.}

The \textbf{contextual similarity} model uses representations induced by the pretrained Clinical BERT model \citep{huang2019clinicalbert} to embed tokens: $f^\mathrm{transformer}_{\boldsymbol{\phi}}: V^{L} \rightarrow \mathbb{R}^{L \times n}$ where $V$ is the set of words in the vocabulary, $L$ is the length of the input text (set of patient records), and $n$ is the dimensionality of the induced embeddings. We take the mean of these vectors to derive a contextual representation. We then take as relevance scores the cosine similarity between the contextual category description representations and each sentence:
\begin{equation}\label{contextual_embedding_mean}
 B_\mathrm{mean}(z) = \mathrm{mean}(f^\mathrm{transformer}_{\boldsymbol{\phi}}(z))
\end{equation}
\begin{equation}
    \boldsymbol{a}_i = \mathrm{Cosine}(B_\mathrm{mean}(\boldsymbol{x}_i), B_\mathrm{mean}(\boldsymbol{d}_q))
\end{equation}

\subsubsection{Distant Supervision}\label{section:distant-supervision}

We now turn to the (distantly) supervised variants that we consider (Figure \ref{fig:model}).
The basic sentence-level attention model we build upon also uses clinical BERT to initialize text encoder weights. However, as this model will be trained, we use the embedding corresponding to the special classification token {\tt [CLS]} (prepended to sentences by BERT) to produce the sentence embedding.\footnote{Note that using this {\tt [CLS]} representation is not advisable in unsupervised settings, hence our use of `mean-pooling' above.} We add a linear layer on top of this to reduce the dimensionality. Here Equation \ref{contextual_embedding_mean} becomes:
\begin{equation}\label{contextual_embedding_cls}
    B_\mathrm{cls}(z) = \boldsymbol{U}_0
    f^\mathrm{transformer}_{\boldsymbol{\phi}}(z)_{\text{\tt [CLS]}} + \boldsymbol{b}_0.
\end{equation}

\noindent Where the ${\tt [CLS]}$ subscript indicates that we extract the embedding from the Clinical BERT output corresponding to the ${\tt [CLS]}$ token.

An embedding for a given query $q$, $\boldsymbol{e}_q$ (discussed in Section \ref{section:query_embeddings}), is then passed to an attention mechanism over sentence embeddings:

\begin{equation}
    \boldsymbol{a}_i = \frac{\exp(B_\mathrm{cls}(\boldsymbol{x}_i) \cdot \boldsymbol{e}_q)}{\sum_{i'} \exp(B_\mathrm{cls}(\boldsymbol{x}_{i'}) \cdot \boldsymbol{e}_q)}.
\end{equation}


\noindent We add two dense layers with a ReLU activation between and sigmoid on top for classification to be trained on top of the concatenation of the final query-specific `global' context vector and the query embedding:
\begin{equation}\label{code_probability}
    P \big(y = 1 \big\vert \boldsymbol{x}, q \big) 
    = 
    \sigma \Big(
    \boldsymbol{U}_2 \:
        \text{ReLU} \Big(
            \boldsymbol{U}_1 \big[
                {\textstyle \sum_i} 
                a_i \, B_\mathrm{cls}(\boldsymbol{x}_i), \boldsymbol{e}_q \big] 
            + \boldsymbol{b}_1 \Big) 
            + \boldsymbol{b}_2 \Big).
\end{equation}
where $y$ is the label that denotes whether or not the patient will experience the condition associated with code $q$ in the future.  For example, if a patient experiences new headaches or blurry vision in a report prior to time-point $t$ (see Section \ref{section:data_extraction}), and after this time-point, they are diagnosed with a brain tumor, the model should predict close to 1 when querying categories like or ancestors of the category ``Malignant Neoplasm of the Brain'', hopefully placing high attention weights on sentences bearing evidence for this. 
This architecture is amenable to training using only the `downstream' label (specifically under a binary cross-entropy loss).
At test time we can then use the induced $\boldsymbol{a}_i$ values as soft relevance scores over individual sentences for a given query.

\subsubsection{Query Embeddings}\label{section:query_embeddings}
The baseline models are not trained, and so the type of embeddings for the diagnosis category being used as the query are limited to TF-IDF embeddings of the category's description, or the raw output of Clinical BERT on it. 

The model discussed in Section \ref{section:distant-supervision} is distantly supervised, and therefore can use more flexible query embeddings. We consider three approaches to form embeddings for these models. One approach simply uses \textbf{indicator embeddings}. However, this type of encoding eliminates any useful signal that might be gleaned from the descriptions associated with the diagnosis category or its position in the hierarchy.

As a second approach we compute \textbf{description embeddings} by applying Eq.~\ref{contextual_embedding_cls} to the natural language description  $\boldsymbol{d}_q$ of the category: $\boldsymbol{e}^\mathrm{description}_q = B_\mathrm{cls}([{\tt [CLS]},\boldsymbol{d}_q])$. This is similar to the Contextual similarity baseline, but replacing $B_\mathrm{mean}$ with $B_\mathrm{cls}$.

The final approach uses the position of a category in the hierarchy to provide structure.  Because the hierarchy is a tree, each category has a unique path from the top node in the hierarchy.  This path is encoded as a concatenation of the natural language descriptions of the nodes along the path separated by {\tt [SEP]} tokens, which is in the vocabulary of special tokens for the BERT tokenizer. Similar to the description embedding approach above, this concatenation of descriptions is then fed through Eq. \ref{contextual_embedding_cls} to create the \textbf{hierarchy embedding}: $\boldsymbol{e}^\mathrm{hierarchy}_q = B_\mathrm{cls}([{\tt [CLS]},\boldsymbol{d}_{\boldsymbol{p}^{(q)}_1},{\tt [SEP]},\boldsymbol{d}_{\boldsymbol{p}^{(q)}_2},...,{\tt [SEP]},\boldsymbol{d}_{\boldsymbol{p}^{(q)}_L}])$ where $\boldsymbol{d}_{\boldsymbol{p}^{(q)}_l}$ represents the $l$th node's description along the path $\boldsymbol{p}^{(q)}$, the path to node $q$ in the hierarchy.\footnote{For both the description embeddings and the hierarchy embeddings, the $B_\mathrm{cls}$ function does share its parameters with the reports encoder.} 


\subsection{Experiments}

We experiment with three types of models as outlined in Section \ref{section:distant-supervision} each of which uses one of the types of query embeddings in Section \ref{section:query_embeddings}: 1) the \textbf{Indicator} model, which uses indicator embedding, 2) the \textbf{Description} model, which uses the description embedding, and 3) the \textbf{Hierarchy} model, which uses the hierarchy embedding.  We train two sets of these models, one on the BWH dataset for 3 epochs and and one on MIMIC-III dataset (which is smaller) for 5 epochs, all with a learning rate of 1e-5 using the Adam optimizer.

We keep all hyper-parameters of clinical BERT the same as in the original paper, initialize all weights to those in clinical BERT, and fine-tune the weights throughout training. 
Clinical BERT outputs embeddings of size 768, and the linear layer in Equation \ref{contextual_embedding_cls} reduces this to 64, which is the size of the hidden vectors for each sentence from the reports and each query embedding.  Therefore, the final prediction layer takes in a context vector of this size and a query embedding of this size.  These two embeddings are concatenated and reduced in the first linear layer of Equation \ref{code_probability} to one vector of size 64.
Because EHR datasets are very imbalanced in the occurance of ICD codes, to prevent naive behavior, we rebalance these as described in Section \ref{section:rebalancing} during training.

\section{Evaluation} 

\subsection{Human (Expert) Labeling}

\begin{figure}
\centering 
\includegraphics[scale=.25]{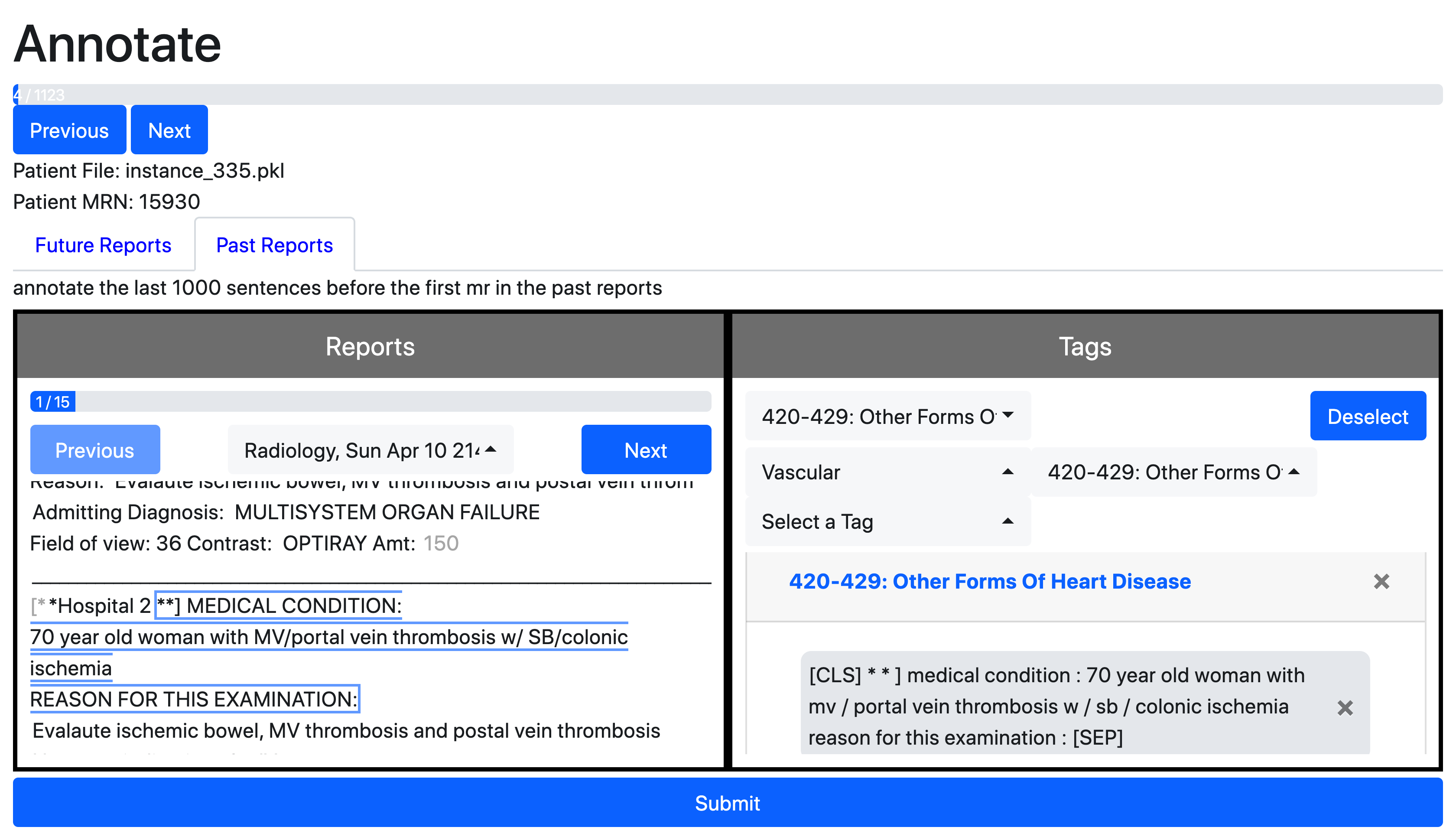}
\caption{Annotation user interface. Clinicians annotate sentences in reports (\emph{left screen}) with category tags (\emph{right screen}).  Data shown is from MIMIC-III.}
\label{fig:interface}
\end{figure}

Recall that our goal is not actually (future) ICD code prediction, but rather extractive summarization that might support particular diagnoses. We do not have direct supervision for this summarization task (which would take the form of $<$\emph{EHR, query/diagnosis, summary}$>$ triplets). We  therefore use the task of predicting downstream diagnoses from clinical notes as a proxy target with which to train extractive summarization models. To evaluate the quality and potential utility of the summaries produced by these models, we perform a direct assessment by domain experts. 

To this end, our radiologist colleagues conduct two rounds of manual annotation. In the first we obtain reference (``gold'') labels that denote whether constituent sentences within the text fields in an EHR should be included in summaries pertaining to a given a query (or diagnosis of interest), also specified by the radiologist. The second annotation exercise entails a \emph{prospective} evaluation of the system; here we ask radiologists to directly assess the subjective quality of model outputs. All annotations are used \textbf{only} for evaluation and not for training of any of the models.

\subsubsection{Reference Standard Summaries}\label{section:reference_standard_summaries}
In this round of annotation, radiologists tag sentences in patient record reports with any diagnosis categories
to which they are relevant (in most cases, none).  We randomly sampled instances from the validation and test sets of the BWH dataset to annotate.

This annotation process employs a set of diagnostic codes with respect to which sentences are to be annotated. 
To enumerate a plausible set, we first asked the radiologists to browse `future reports' (going forward up to one year past $t$ as discussed in Section \ref{section:data_extraction}) and tagging these reports with clinically relevant diagnoses. 
This step yielded a summary of all salient conditions that appear in the patient's `future'. This serves two purposes. First, it allows us to validate that the diagnosis categories used as targets for that instance during training (derived from the ICD codes present in the EHR) were valid, and to expand on them. ICD codes serve as noisy proxies for diagnoses. We therefore report agreement between the annotated future diagnosis categories and the ICD code based diagnosis categories (see Table \ref{table:future_annotation_statistics}). Second, and more importantly, this strategy allowed the radiologists to identify categories of interest to use when tagging sentences in the previous medical record.

The radiologists used the manually created hierarchy (see Section \ref{section:data_extraction}) as a starting point from which to tag both future reports and past reports and were also allowed to add new categories to the hierarchy.  
These ``custom'' categories can still be used to query models that condition extraction on natural language (rather than categorical) queries.\footnote{This includes both of the baseline models as well as the trained models that use the description embedding or hierarchy embedding (both of which are natural language based).} For the one model that does not condition on natural language --- the indicator model --- we simply eliminate from consideration annotations that use custom tags.

We use the reference label summaries collected during this round of annotations to compute precision-recall curves over sentence percentile thresholds, where the percentile is the fraction of total unique sentences ranked above a given sentence by the model. Here the `true positives' are sentences that were extracted by the model given the threshold and present in the reference summary; `false positives' are sentences that were extracted by the model and not present in the reference summary; and `false negatives' are sentences present in the reference summary that were not extracted by the model.

Two radiologists perform this round of annotations, so we compute annotator agreement over 4 instances (see Table \ref{table:annotator_agreement}).  Though these annotators agree on a number of queries, Annotator 2 clearly marks more queries and sentences for each query than Annotator 1.  Because these annotations are subjective and should be used to evaluate how well the models do with respect to what {\it radiologists} find important, we deliberately give no specific guidelines to the annotators regarding the number of sentences to annotate or the number of queries for which to annotate sentences.

\subsubsection{Model Validation}\label{section:model_validation}
We also consider a more subjective, but more direct and explicit, assessment of summary quality. Here we ask radiologists to mark whether each sentence comprising summaries produced by a particular model was relevant to the diagnosis category (the query) upon which summary extraction was conditioned. For this round of annotations, we sample a fresh set of instances separate from those used to create the reference standard summaries.

As in Section \ref{section:reference_standard_summaries}, radiologists first browse future reports to select the queries on which to test the model.  In practice, the radiologist would not have access to the future but will have access to the MR images.  These images may or may not tell the radiologist about salient diagnoses that will later be apparent in the patient. The reason that we provide the radiologist with more information than they would have in practice, is that this allows them to test the model on accurate diagnoses that may not normally be considered at the time of the MR, allowing for richer annotations.   
Due to the time cost of annotation, we validate only a subset of the models, comparing the TF-IDF similarity (the highest-performing baseline) and the Hierarchy model (the highest-performing distantly supervised model from the BWH dataset models).

For these models, we calculate a \textbf{validated precision} score which is simply the fraction of sentences marked as relevant by a radiologist out of the total number of sentences extracted by the model across all instances and categories annotated.



\subsection{Results}

\begin{figure}
\centering 
\includegraphics[scale=.6]{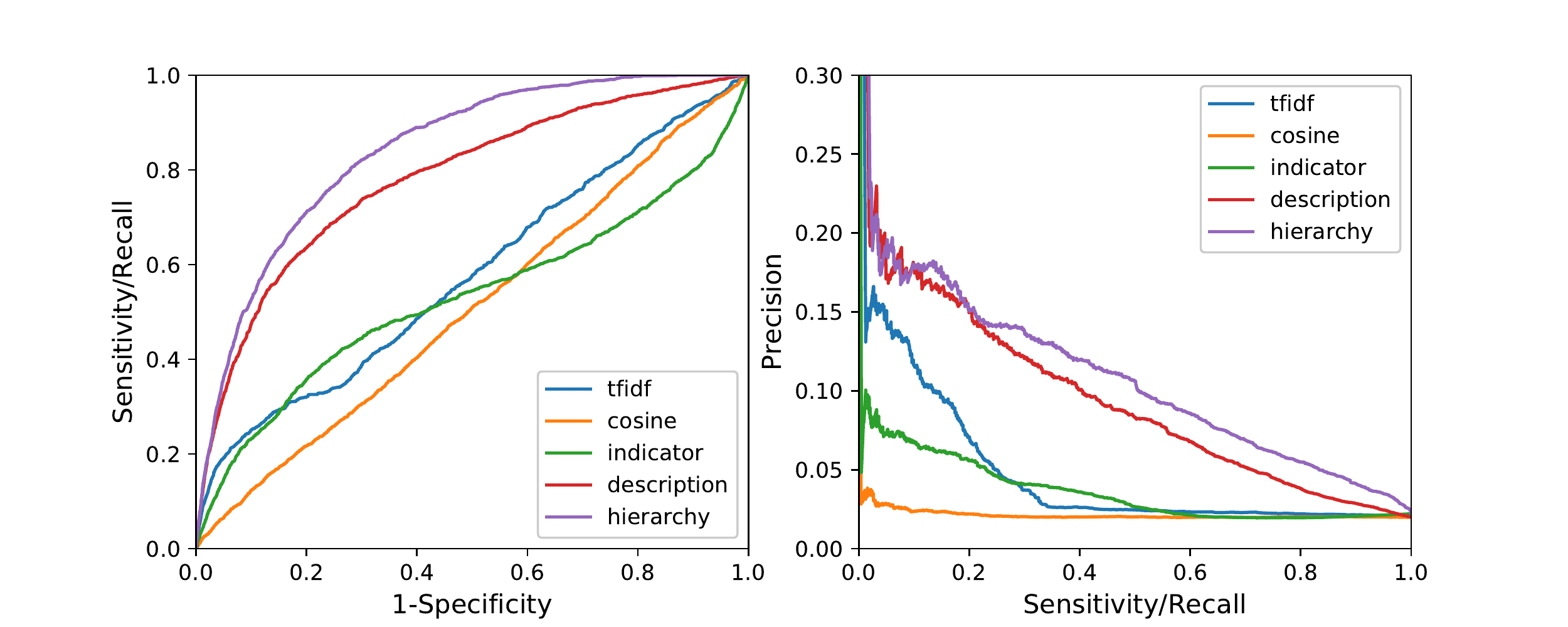}
\caption{Reference summary \textbf{sentence retrieval}. We compute Receiver Operator Characteristic (ROC) curve (left) and Precision Recall (PR) curve (right) over thresholds on the percentile of a sentence in the reports as determined by each model (micro-averaging over summaries).}
\label{fig:roc_prc}
\end{figure}

\begin{table}
\footnotesize
\begin{center}
\begin{tabular}{ l c c c c c c c }
    \hline
     & \multicolumn{6}{c}{Reference Summaries} & Model Validation \\
    \cline{2-8}
     & \multirow{2}{*}{AUROC} & \multirow{2}{*}{Avg. P} &  \multirow{2}{*}{Avg. NDCG} & \multicolumn{3}{c}{top 20} & \multirow{2}{*}{\makecell{Validated P\\ (top 20)}} \\
     & & & & P & R & F1 & \\
    \hline
    TF-IDF Similarity & .578 & .045 & .407 & .085 & .163 & .112 & 0.173 \\
    Cosine Similarity & .508 & .021 & .279 & .029 & .055 & .038 & - \\
    \hline
    Indicator & .531 & .036 & .350 & .074 & .126 & .093 & - \\
    Description & .782 & .093 & .425 & .137 & .261 & .179 & - \\
    Hierarchy & \textbf{.842} & \textbf{.106} & .447 & \textbf{.144} & .275 & .189 & \textbf{0.186} \\
    \hline
    Indicator (MIMIC) & .540 & .043 & .366 & .082 & .141 & .104 & - \\
    Description (MIMIC) & .783 & .104 & \textbf{.481} & .146 & \textbf{.278} & \textbf{.192} & - \\
    Hierarchy (MIMIC) & .760 & .085 & .411 & .117 & .223 & .153 & - \\
    \hline
\end{tabular}
\end{center}
\caption{Extractive summarization results on the BWH dataset. AUC and Average precision are computed for the ROC and PR curves in Figure \ref{fig:roc_prc} respectively.  The Normalized Discounted Cumulative Gain (NDCG) is calculated for each query's results (using the reference summary sentences) and averaged over all the queries.  To compare with validated precision, we compute micro-averaged Precision, Recall, and F1 (using the reference summary sentences) at a top 20 sentence threshold as ranked by each model for each query.  We also compute the micro-averaged validated precision using Model Validation annotations over the top 20 sentences (see Section \ref{section:model_validation}). The bottom three rows correspond to cases in which we train on the MIMIC dataset and \emph{transfer} these models to the BWH dataset.}
\label{table:results}
\end{table}

Table \ref{table:results} shows that the hierarchy model performs better than the other models trained on the BWH dataset and the baseline both in retrospective (Section \ref{section:reference_standard_summaries}) and prospective (Section \ref{section:model_validation}) evaluations, with the description model following not far behind.  We emphasize that while these models are `trained', they use \emph{only} distant supervision that is available effectively for free (modulo a domain expert enumerating relevant ICD code groups). Hence we believe the direct comparison to unsupervised baselines is warranted. 

We can see from the PR curve in Figure \ref{fig:roc_prc} that about 20 percent of the highest-ranked sentences are relevant, which is a 10-fold improvement over the precision of the whole document (yielding about 2 percent relevant sentences). Given that it seems radiologists rarely have time to look at the medical record at all at present, even this preliminary increase in precision might prove invaluable for at least some cases. 

The TF-IDF model still performs surprisingly well given that the task is predicting evidence for as-yet \emph{undiagnosed codes}. TF-IDF yields non-zero similarity scores for 361 out of 1840 reference sentences annotated in total (recounting sentences tagged under more than one query). For these, we find that subjectively, the annotated sentences fall into two categories: (1) the patient {\it had} already been diagnosed with the diagnosis, or; (2) a list of differential diagnostic possibilities had been proposed, but no definitive diagnosis had been selected. Though instances were formed such that there should be at least one salient diagnosis that has not been made yet, it does not ensure that every salient diagnosis of the patient is undiagnosed at time-point $t$. For these cases, radiologists were permitted to annotate the diagnosing sentence in the past reports with the corresponding tag, which is likely to have text overlap with the sentence. It is useful to identify relevant pre-existing diagnoses to radiologists because in many cases, this information is not available a priori.  However, in the future, it may be prudent to create a subset of annotations from which to instruct annotators to remove diagnoses that they find have already been made prior to time-point $t$ to explicitly evaluate models on only subtle evidence of a certain diagnosis.
Unfortunately, due to the confidentiality of the data, we cannot present examples of this. However, we do present plots in the Appendix (Figures \ref{fig:roc_prc_tfidf_zero}) that show that for reference sentences that have zero TF-IDF similarity scores, our proposed models still do well.

We compute the human validated precision using the prospectively collected annotations of the top 20 sentences ranked by the models for each query.  We micro-average the precision, calculating it as the total number of sentences marked relevant divided by the total number of sentences reviewed.  In Table \ref{table:results}, the precision for retrieving reference summary sentences in the top 20 sentences ranked by models is similar to the validated precision.  It is worth noting that we expect human validated precision to be higher because of confirmation bias.

One of the most surprising details about Table \ref{table:results} is how well the models that we train on MIMIC-III and then transfer to the BWH dataset fare, even outperforming the best model (the hierarchy model) in terms of a few metrics. We plan to investigate this finding in more detail in future work, but view evidence of such transfer as promising: This suggests models trained on data from one EHR may be deployed elsewhere and still provide meaningful extractive summaries. We provide additional plots with the MIMIC-III models in the Appendix (Figures \ref{fig:roc_prc_mimic} and \ref{fig:roc_prc_code_prediction_mimic}).

\section{Related Work} 
\label{section:related-work} 

We briefly review related work in the subareas to which this effort is related. 

\paragraph{Machine Learning for Radiology.} There is a long history of work applying machine learning methods in radiology \citep{wang2012machine}. Such technologies are commonly used for computer aided diagnosis of medical images \citep{doi2007computer}. Some recent work has suggested that for certain specific tasks (e.g., diagnosing pneumonia on the basis of a chest x-ray), modern machine learning models can in fact \emph{outperform} radiologists \citep{rajpurkar2018deep}.\footnote{It should be noted that this comparison was under extremely constrained settings \citep{larvie2019machine}.} In real clinical settings, practical use of such technologies will likely be via aiding, rather than replacing, radiologists.
	

\paragraph{Content-Based Medical Information Retrieval for Radiology.} There is a substantial body of work on image retrieval methods for diagnostic medical imaging \citep{kalpathy2015evaluating,demner2009annotation,li2018large}. Such approaches accept a query image and retrieve similar cases from a large database of previously assessed images which may inform diagnosis \citep{akgul2011content}. These efforts have nearly exclusively focused on retrieval based on imaging, ignoring unstructured notes associated with patients. The few investigations that have considered text have done so in the context of ``multi-modal'' retrieval, where the aim is still to retrieve images similar to a query image, but to do so also informed by additional query text \citep{cao2014medical,neveol2009natural,kumar2013content}. The approach considered here, which focusses on retrieving and summarizing relevant notes from a patient history, is thus complementary to these approaches.

\paragraph{Text Summarization.} Methods for general text summarization have been studied in the NLP research community for decades \citep{maybury1999advances}. Broadly, models for text summarization can be grouped into extractive and abstractive techniques. Models of the former variety extract snippets from sources verbatim to compose summaries, while those of the latter type generate novel summary text. Recently, neural models have engendered rapid progress on general text summarization tasks \citep{chopra2016abstractive,nallapati2017summarunner,cao2016attsum,see2017get}. Our model (Section \ref{section:models}) for extractive, query-focused summarization builds upon these prior efforts. However, here we assume access to only \emph{distant} supervision.


\paragraph{Summarizing EHR.} More specific to the present work, the task of summarizing EHR data has received considerable research attention over the years, dating back to the 1970s \citep{rogers1979impact,simborg1981summary}. We do not attempt a general survey of such approaches here, but instead point the reader to Pivovarov and Elhadad’s relatively recent, exhaustive survey \citep{pivovarov2015automated}, which reviews and contrasts 12 published methods for summarizing EHR.

\paragraph{Predicting ICD Codes From Clinical Notes} Finally, we note that considerable effort has been invested in designing models that can automatically tag clinical notes in EHR with relevant ICD codes, dating back to the mid-90s \citep{larkey1995automatic}. This work, however, is only tangentially related to our approach, in which we use collections of \emph{future} ICD codes as proxy targets to induce distant supervision. A recent related effort in this line of work also used the idea of extracting snippets relevant to individual codes \citep{mullenbach-etal-2018-explainable}, although the focus remained primarily on code prediction.  Our work is also distinct from this in that the model predicts codes from future time-points relative to the input text rather than codes aligned with the text.


\paragraph{NLP in Radiology.} Finally, there has also been some prior work applying NLP to radiology; a recent article provides a systematic review of these methods and applications \citep{pons2016natural}. The vast majority of these efforts have focused on processing and summarizing \emph{radiology reports} \citep{sinha2000interactive}. This remains an active area of research \citep{peng2018negbio,zhang2018learning,zhang2019optimizing}. There have also been some efforts to use NLP in radiology that aim to provide clinical support. However, these have primarily been concerned with identifying transcription errors introduced by speech recognition systems \citep{voll2008improving} and automated coding of reports \citep{farkas2008automatic}. In contrast to these prior applications of NLP in radiology, which focus on processing existing radiology reports, here we have proposed models intended to aid the radiologist in writing reports in the first place by easing access to the relevant contextualizing information hidden in the EHR.

\section{Discussion}

We have proposed the task of inducing conditional extractive summaries of notes within electronic health records (EHR) to aid clinicians (here, radiologists). We have designed and evaluated models for that that use only groups of diagnosis codes as (distant) supervision. These models use Clinical BERT \cite{huang2019clinicalbert} to jointly encode queries (e.g., ICD code descriptions) and notes, and then predict (groups of) ICD codes corresponding to potential diagnoses of interest. The latter is only a proxy task used to train the model; we are not actually interested in these code predictions. Rather, we use the top-level attention distributions as a scoring mechanism to retrieve potentially relevant snippets, obviating the need for explicit supervision regarding snippet relevance.

We developed an interface to allow radiologists to annotate snippets within notes with respect to their relevance to clinical queries, i.e., downstream diagnoses. We also asked them to directly assess the relevance of different model outputs for given queries, without revealing which models produced which outputs.



Our results demonstrate that the proposed distantly supervised models significantly outperform unsupervised baselines. As far as we are aware, this is the first attempt to conditionally summarize medical records without direct supervision. Because these models exploit only distant supervision, they can in principle be applied widely, beyond radiology, to other specialities. 

In future work, we would like to assess the degree to which incorporating a small amount of \emph{direct} supervision might improve model performance \citep{wallace2016extracting}. Eventually, we aim to actually evaluate the utility of this system when used in clinical practice initially specifically in neuroradiology, later in other radiology sub-specialties and ultimately in a more generally usable form.




\bibliography{references}

\appendix

\renewcommand{\thefigure}{A\arabic{figure}}
\setcounter{figure}{0}

\renewcommand{\thetable}{A\arabic{table}}
\setcounter{table}{0}

\section{Manually Created Hierarchy}

Here we discuss the manually created hierarchy from Section \ref{section:data_extraction} in more detail. One of our radiologist collaborators constructed this hierarchy of diagnosis categories specifically of interest to radiologists based on ICD-9 Diagnosis codes \citep{ICD9}. The top level categories are Trauma, Infection, Neoplasm, Demyelinating/Autoimmune, Neurodegenerative, and Metabolic and Endocrine. The tree has varying depths in different parts, but each leaf node is either a collection of codes (e.g., 420-429) or one code (e.g., 432.0). We extend this by collecting the ICD-9 code(s) that correspond to the leaf node, mapping these to ICD-10-CM \citep{ICD10} codes using a general equivalence mapping \citep{ICD9to10}, and adding these nodes as children of the category. We use ICD-10 for these leaf nodes because it is more granular. During dataset extraction, we map both ICD-10-CM codes and ICD-9 codes (using the same mapping as before) to their corresponding leaf nodes. The final hierarchy has 3,310 nodes in it and a maximum depth depth of 6.

Our radiologist also annotated descriptions along with each of the categories excluding the top-level ones (i.e. 420-429 has a description ``Other Forms Of Heart Disease'').  For the top-level categories, we use the node name as the description (e.g., ``Trauma''), and for the ICD-10-CM leaf nodes, we use the descriptions already available.

\section{ICD Code Re-balancing}\label{section:rebalancing}

EHR datasets are unbalanced in the occurrence of ICD codes in two ways: 1) each ICD code occurs rarely and 2) the rarity of each ICD code is not equal.  This means that when making a binary prediction for a specific ICD code or corresponding category, the first-order naive behavior would be to only predict 0 (negative).  If one rebalances to eliminate this prediction, a second-order naive behavior would emerge: the model's prediction would depend only on the code or category and not the text.  To eliminate the first behavior, we weight the loss on negatively labeled categories for batch $b$ by 
\begin{equation}
    \frac{\textrm{neg}^{(\textrm{batch})}_b}{\textrm{pos}^{(\textrm{batch})}_b}
\end{equation}
where $\textrm{neg}^{(\textrm{batch})
}_b$ is the number of negative labels in the batch and $\textrm{pos}^{(\textrm{batch})}_b$ is the number of positive labels in the batch.  We then perform a more traditional rebalancing to eliminate the second behavior by resampling the negative categories to have the same distribution as the positive categories. To form the new negative categories for an instance, we sample (with replacement) $n$ categories from a distribution over the original negative categories of an instance that weights each category $c$ by 
\begin{equation}
    \left(\frac{\textrm{pos}^{(\textrm{train})}_c}{\sum_{c'} \textrm{pos}^{(\textrm{train})}_{c'}}\right) \frac{1}{\textrm{neg}^{(\textrm{train})}_c}
\end{equation}
where $\textrm{pos}^{(\textrm{train})}_c$ and $\textrm{neg}^{(\textrm{train})}_c$ are the number of positive and negative occurrences, respectively, of category $c$ in the training data.  In general, the number of samples $n$ can be the number of original negative categories in the instance, but for computational efficiency, we downsample and only keep each of these with probability $p$ chosen to be $.01$, so $n \sim \textrm{Binomial}(\textrm{neg}^{(\textrm{instance})}_i, p)$ where $\textrm{neg}^{(\textrm{instance})}_i$ is the number of negative categories originally present in the instance $i$.

\section{Additional Summarization Results}

\begin{figure}[H]
\centering 
\includegraphics[scale=.6]{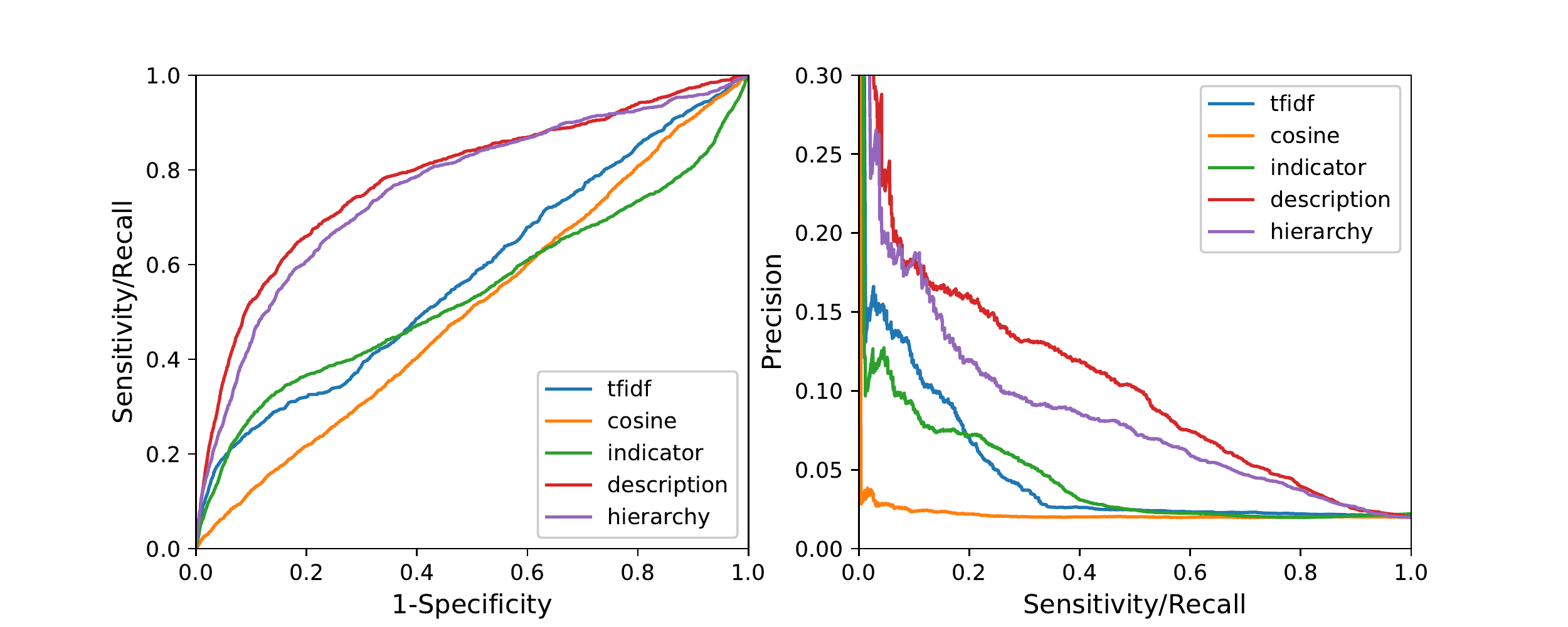}
\caption{Same as Figure \ref{fig:roc_prc} but the distantly supervised models are trained on MIMIC.  Results still show performance of these models on {\it BWH} dataset annotations.}
\label{fig:roc_prc_mimic}
\end{figure}

Here we share a more fine-grained analysis of the summarization results. Figure \ref{fig:roc_prc_mimic} provides curves for the models trained on MIMIC instead of the BWH dataset. Because annotations were only collected on the BWH dataset, we can only test on this set. Therefore, to reiterate our observation from the Results section, it is surprising that the transfer is so successful. It is also interesting that in this set of models, the Description model performs better than the Hierarchy model, especially given that the hierarchy model has strictly more information.  It is unclear whether this is random or if there is some systematic reason for this.  It may also simply be evidence that these models should be trained for longer, though it is difficult to tell how long by looking at the loss function, especially because it does not directly correspond to our ultimate goal.

\begin{figure}[H]
\centering 
\includegraphics[scale=.6]{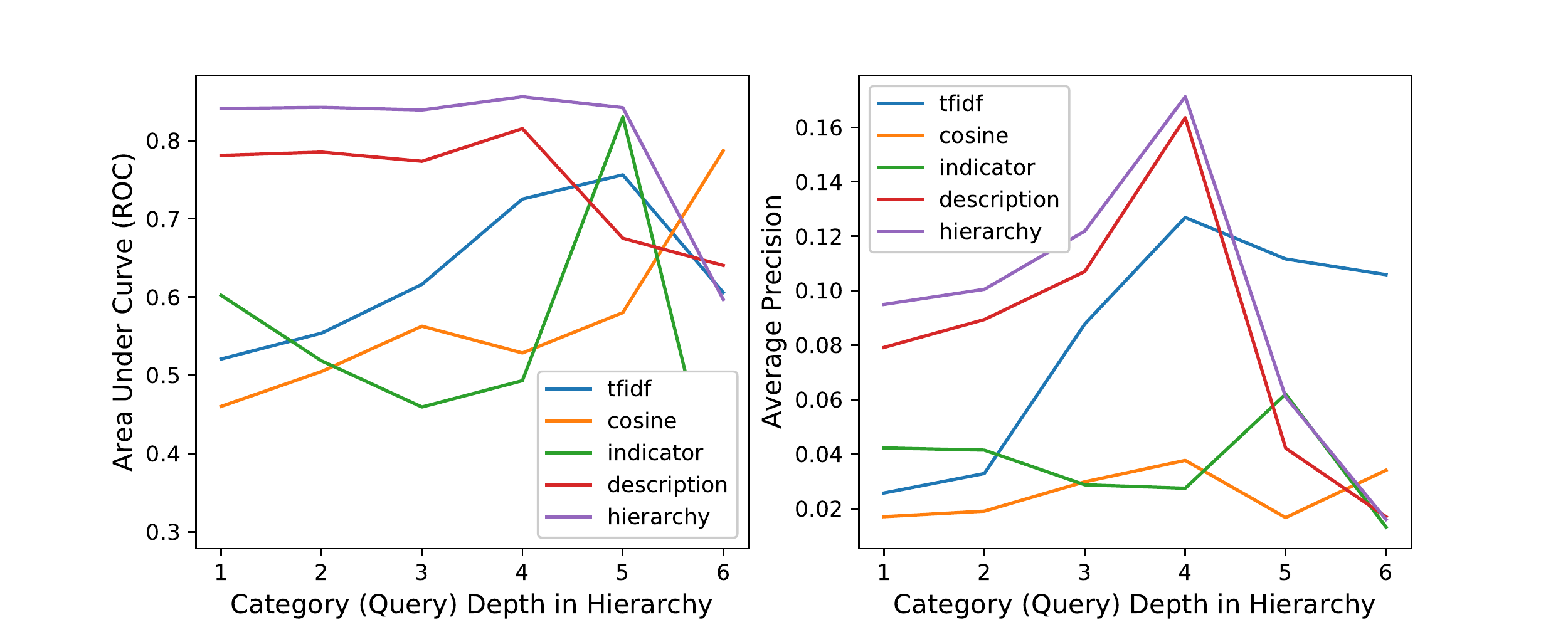}
\caption{This shows AUROC (left) and Average Precision (right) on reference sentence retrieval for queries at each depth in isolation.}
\label{fig:rocauc_avgprc_by_depth}
\end{figure}

In order to get a sense of how models perform on queries at each depth of the hierarchy, we plot AUROC and Average Precision for these queries in isolation (Figure \ref{fig:rocauc_avgprc_by_depth}).  We can see that models have a spike at either 4 or 5 across these metrics in the hierarchy.  For reference, the number of codes at each depth is 7, 87, 2455, 689, 57, and 15 respectively.  The last three depths are almost all leaf nodes, and the last two depths have very little numbers of codes, so these are probably fairly noisy.  As the categories get more specific, there is less training data on them but they also can capture more explicitly what the relevant evidence is.  Especially in the average precision graph, it is evident that at least over the first 4 depths, there is generally an increase in precision which we expect from this increase in specificity.

\begin{figure}[H]
\centering 
\includegraphics[scale=.6]{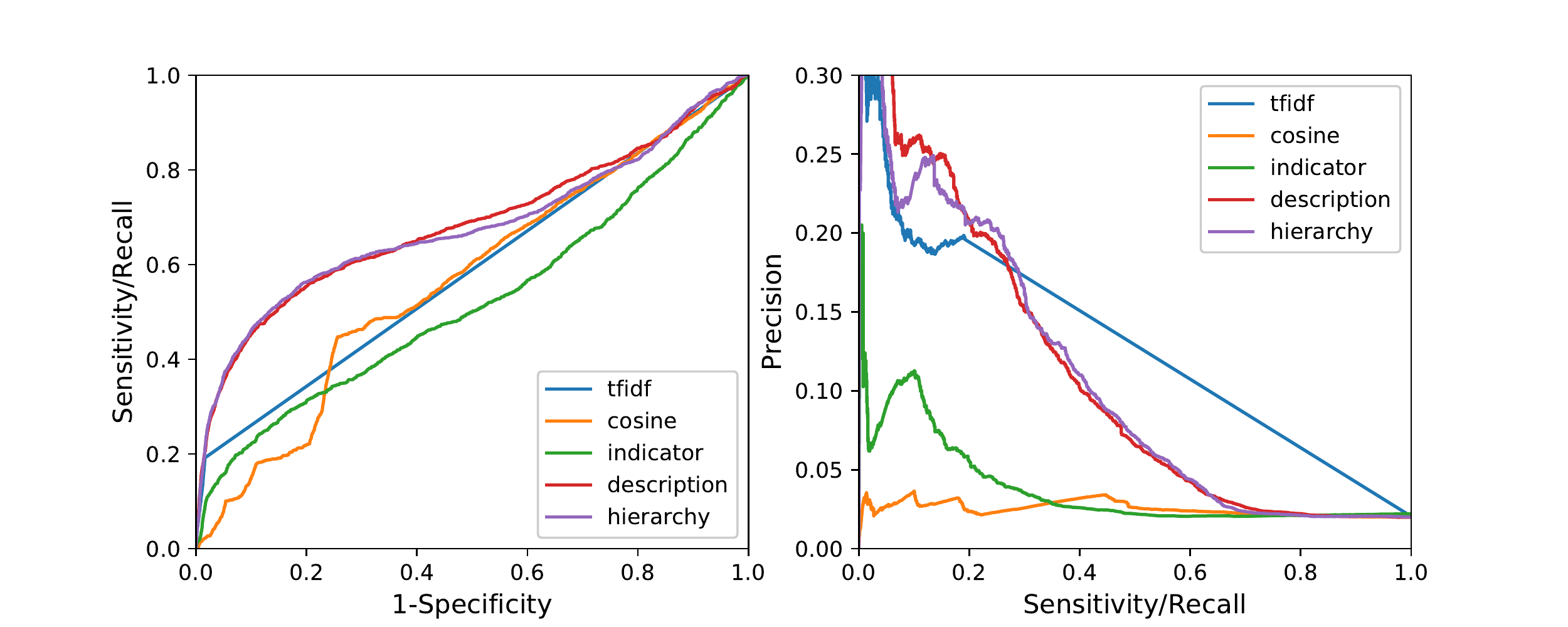}
\caption{This shows ROC and PR curves for sentence retrieval over thresholds based on the attention given to sentences rather than their percentile (as in Figure \ref{fig:roc_prc}).}
\label{fig:roc_prc_attention}
\end{figure}

We also show ROC and PR curves using thresholding based on the attention (or similarity score in the case of the baselines) on a sentence rather than the precentile of a sentence (Figure \ref{fig:roc_prc_attention}).  This means that there will likely be more variance in the number of sentences included in the summary on each query.  It is expected that this will benefit the TF-IDF similarity model in precision because using percentile thresholds, there are many sentences which have a zero TF-IDF similarity score that will be included in the TF-IDF summary because not very many sentences have a non-zero TF-IDF score.  Therefore, when those are not included using the attention thresholding, precision will jump.  However, it also means that there is no threshold that will achieve anywhere in between around .2 recall and total recall.  This is why we see a straight line in that range.

\begin{figure}[H]
\centering 
\includegraphics[scale=.6]{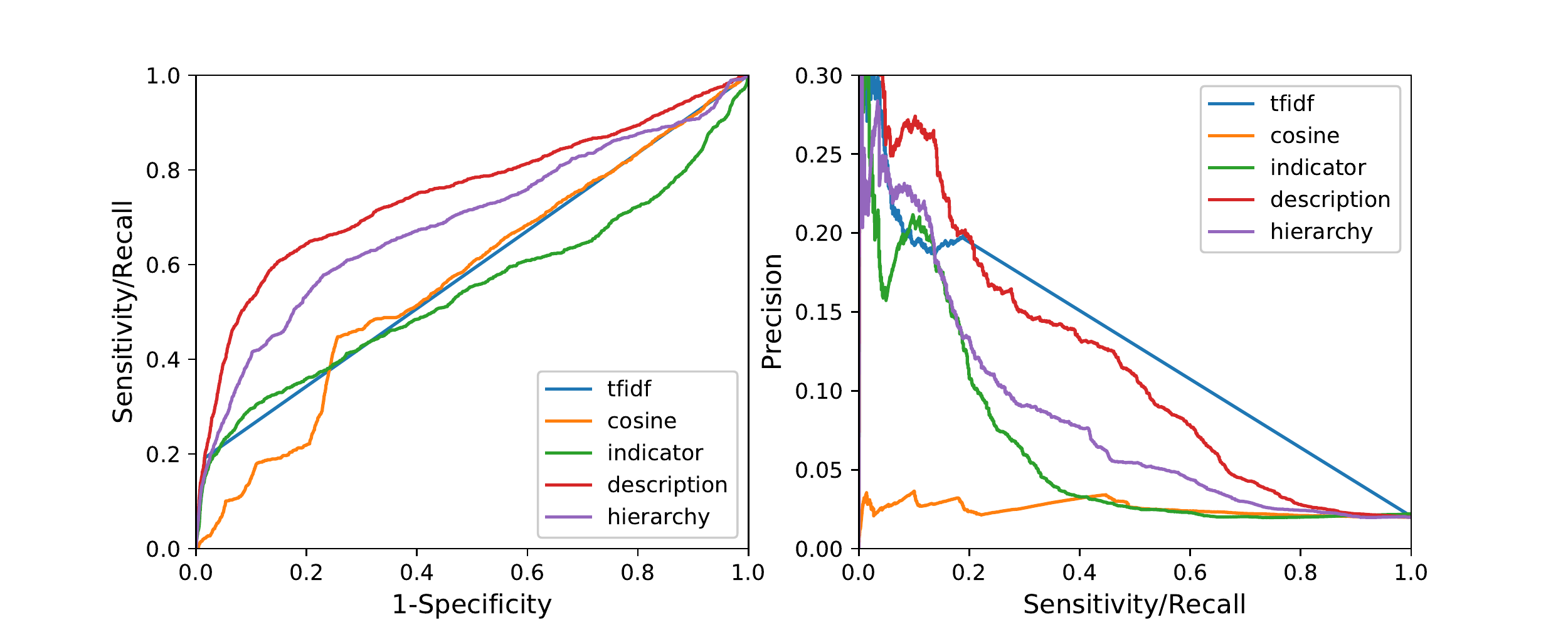}
\caption{Same as Figure \ref{fig:roc_prc_attention} for MIMIC models.}
\label{fig:roc_prc_attention_mimic}
\end{figure}

\begin{table}
\small
\begin{center}
\begin{tabular}{ l c c }
    \hline
     & AUROC & Avg. P\\
    \hline
    TF-IDF Similarity & .588 & .061 \\
    Cosine Similarity & .569 & .025 \\
    \hline
    Indicator & .523 & .038 \\
    Description & .684 & .116 \\
    Hierarchy & .679 & .110 \\
    \hline
    Indicator (MIMIC) & .546 & .065 \\
    Description (MIMIC) & \textbf{.751} & \textbf{.121} \\
    Hierarchy (MIMIC) & .688 & .083 \\
    \hline
\end{tabular}
\end{center}
\caption{AUROC and Average Precision for curves in Figures \ref{fig:roc_prc_attention} and \ref{fig:roc_prc_attention_mimic}.}
\label{table:results_attention}
\end{table}

\begin{figure}[H]
\centering 
\includegraphics[scale=.6]{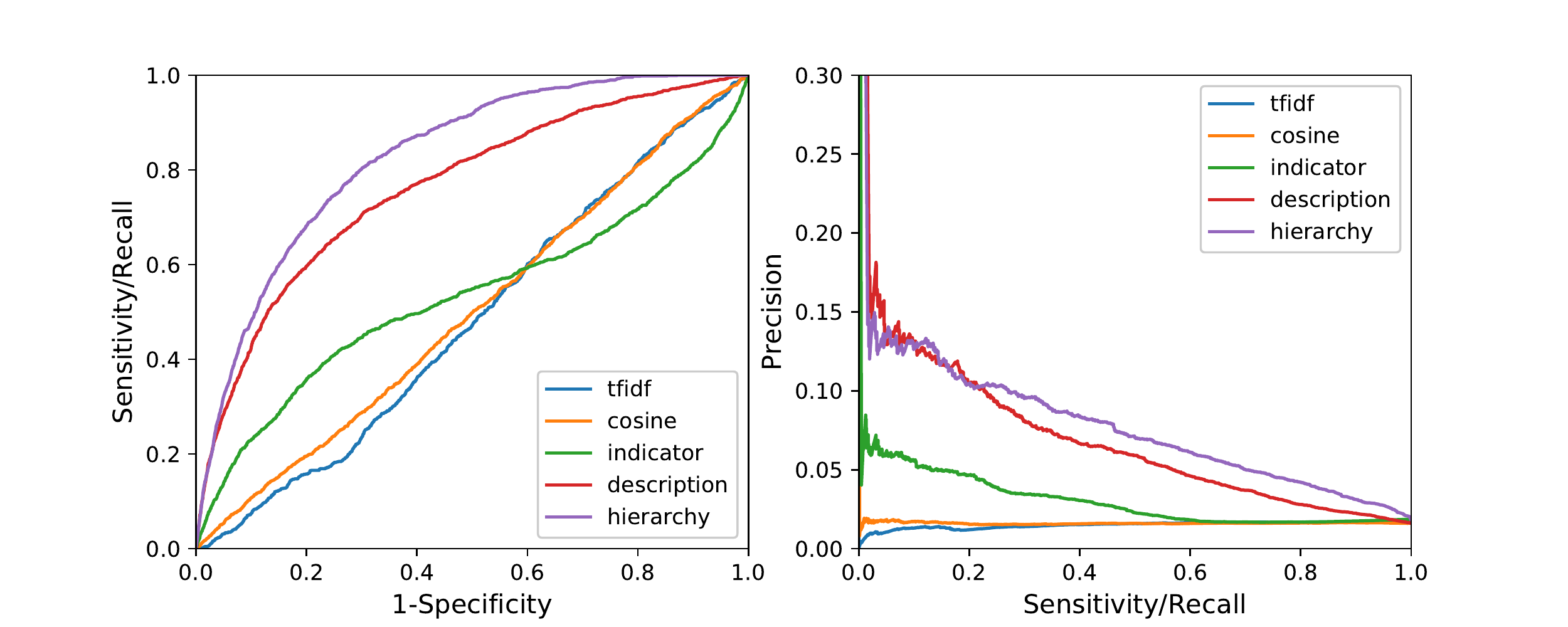}
\caption{This shows the same plots in Figure \ref{fig:roc_prc} but excluding all of the reference sentences where the TF-IDF has a non-zero score, demonstrating that our models perform well on ``subtle'' evidence.}
\label{fig:roc_prc_tfidf_zero}
\end{figure}

In Figure \ref{fig:roc_prc_tfidf_zero}, we show that our models still preform well when excluding all annotations that have a zero TF-IDF similarity score.  Though there is some decrease in performance, this shows that these models are retreiving ``subtle'' evidence that would not be able to be captured by any sort of word matching search.

\begin{figure}[H]
\centering 
\includegraphics[scale=.6]{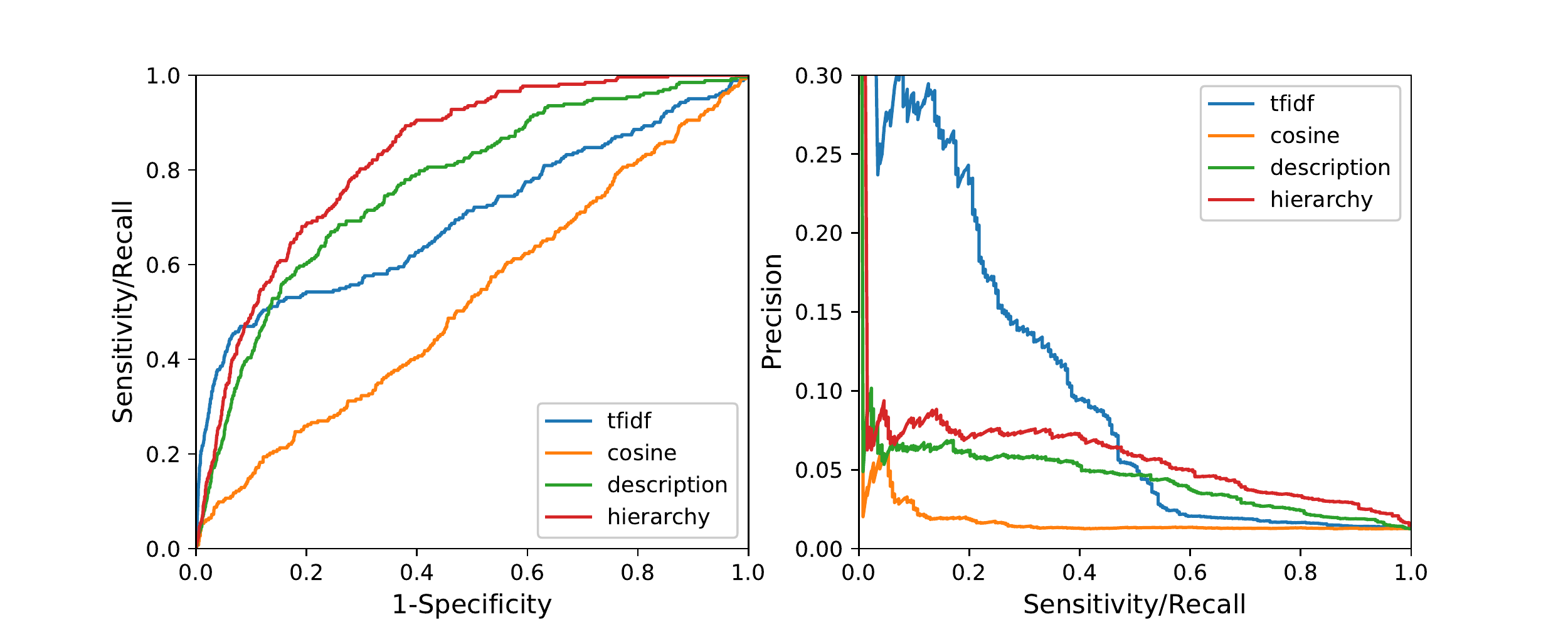}
\caption{Same as Figure \ref{fig:roc_prc} but limiting to only custom queries.}
\label{fig:roc_prc_custom}
\end{figure}

\begin{figure}[H]
\centering
\includegraphics[scale=.6]{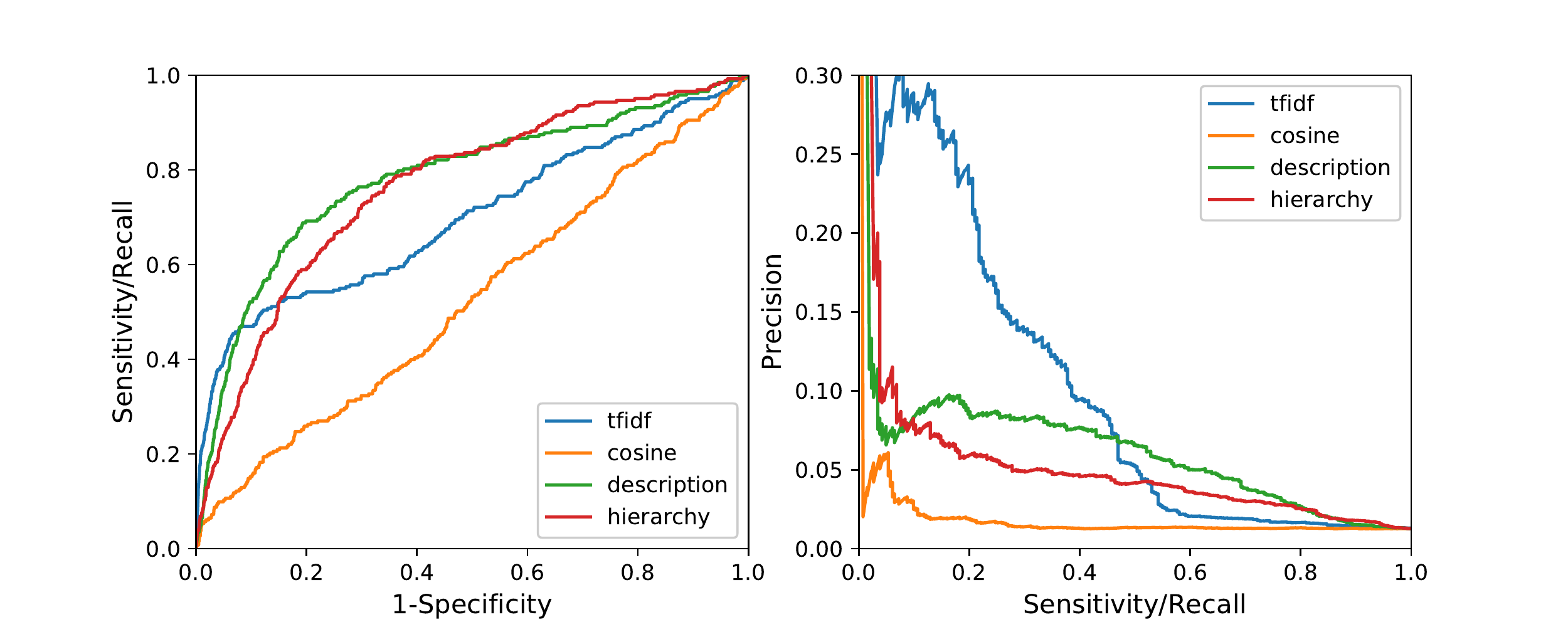}
\caption{Same as Figure \ref{fig:roc_prc_mimic} but limiting to only custom queries.}
\label{fig:roc_prc_custom_mimic}
\end{figure}

Because, for 4 of the models (both baselines and the description and hierarchy models), new ``custom'' queries not present in the training data are allowed (as discussed in Section \ref{section:reference_standard_summaries}), we isolate these annotations and show performance on just this subset (Figures \ref{fig:roc_prc_custom} and \ref{fig:roc_prc_custom_mimic} and Table \ref{table:results_custom}).  TF-IDF clearly now dominates, but this is because when annotators do not find a relevant annotation, they are much more likely to create one that relates explicitly to the sentence they are trying to annotate.  The annotator in model validation (see Section \ref{section:model_validation}) chose not to use custom queries on the small set of instances annotated.  We expect that had the annotator actually performed custom queries in the model validation annotations, the results would be more similar to the overall results because the annotator would not be looking at the evidence when proposing the query.  We hope to investigate this further in future work.

\begin{table}
\footnotesize
\begin{center}
\begin{tabular}{ l c c c c c c }
    \hline
     & \multicolumn{6}{c}{Reference Summaries}\\
    \cline{2-7}
     & \multirow{2}{*}{AUROC} & \multirow{2}{*}{Avg. P} &  \multirow{2}{*}{Avg. NDCG} & \multicolumn{3}{c}{top 20}\\
     & & & & P & R & F1\\
    \hline
    TF-IDF Similarity & .699 & \textbf{.100} & \textbf{.511}  & \textbf{.110} & \textbf{.336} & \textbf{.166} \\
    Cosine Similarity & .524 & .017 & .253 & .029 & .087 & .043 \\
    \hline
    Description & .768 & .044 & .319 & .060 & .183 & .090 \\
    Hierarchy & \textbf{.831} & .056 & .338 & .069 & .209 & .104 \\
    \hline
    Description (MIMIC) & .783 & .059 & .375 & .079 & .240 & .119 \\
    Hierarchy (MIMIC) & .763 & .048 & .351 & .059 & .179 & .089 \\
    \hline
\end{tabular}
\end{center}
\caption{Same as Table \ref{table:results} but limiting to only custom queries.  In the annotation examples for Model Validation, the annotator did not use any custom queries, which is why this column is omitted.}
\label{table:results_custom}
\end{table}

\section{Code Prediction Results}

\begin{figure}[H]
\centering 
\includegraphics[scale=.6]{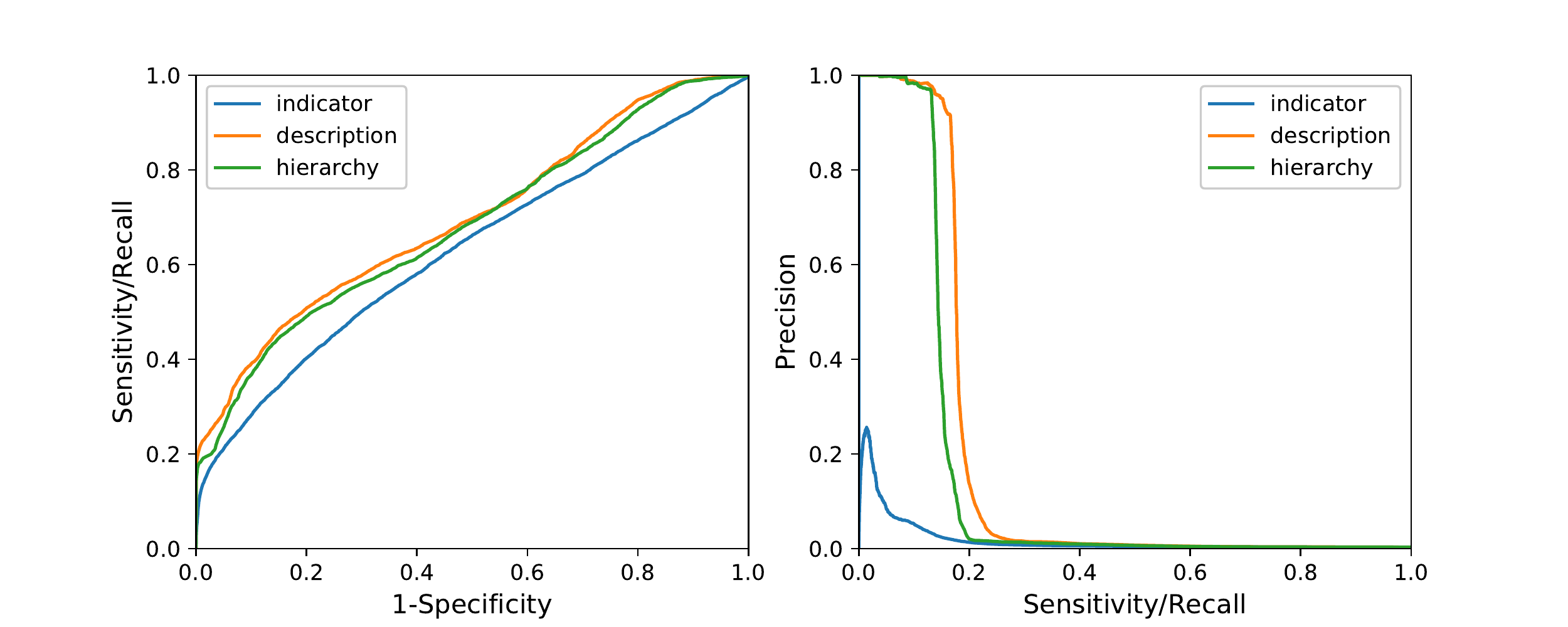}
\caption{This shows ROC and PR curves for code prediction.}
\label{fig:roc_prc_code_prediction}
\end{figure}

\begin{figure}[H]
\centering 
\includegraphics[scale=.6]{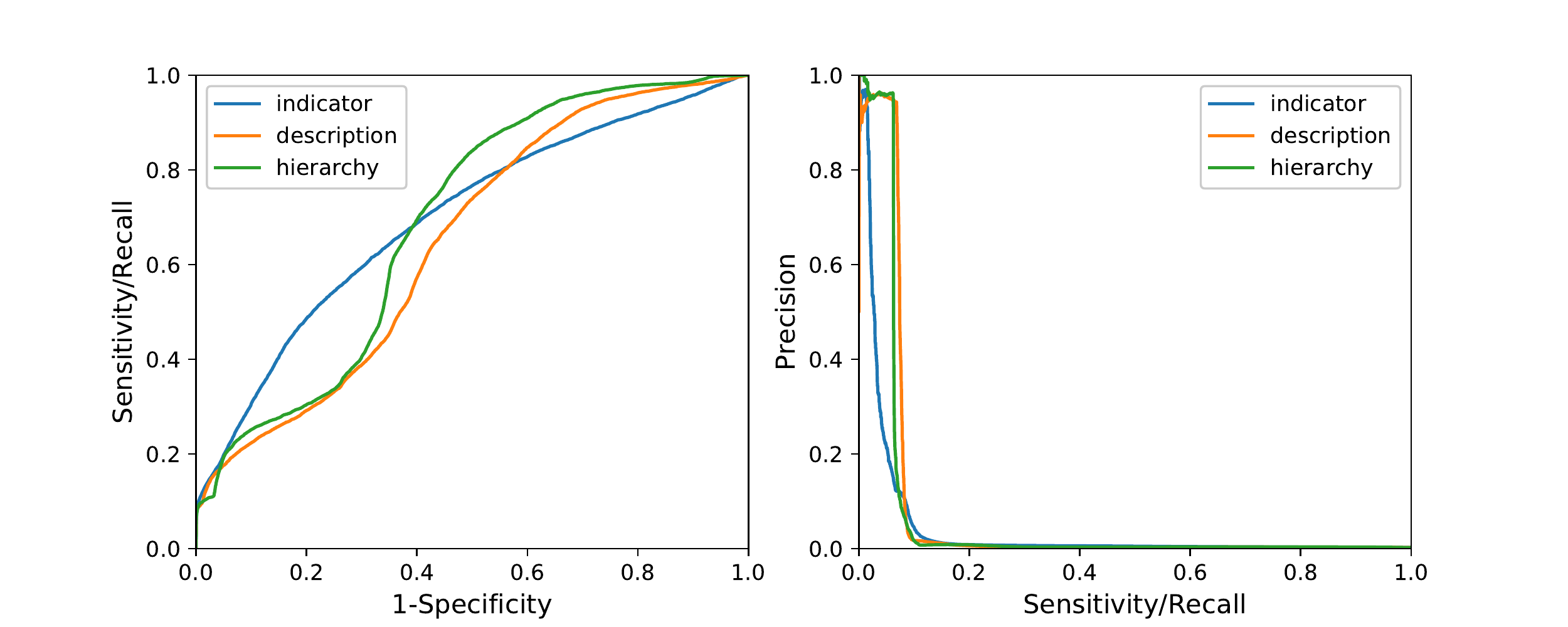}
\caption{Same as Figure \ref{fig:roc_prc_code_prediction} but the models are trained on MIMIC and performance is measured on MIMIC.}
\label{fig:roc_prc_code_prediction_mimic}
\end{figure}

Though this is not the main focus of the paper, we do show ROC and PR curves for code prediction in Figures \ref{fig:roc_prc_code_prediction} and \ref{fig:roc_prc_code_prediction_mimic}.  Note that these should not be compared to results in papers such as \cite{mullenbach-etal-2018-explainable} because our model predicts future codes, not those aligned with the input text (see Section \ref{section:related-work}).

\begin{table}
\small
\begin{center}
\begin{tabular}{ l c c }
    \hline
     & AUROC & Avg. P\\
    \hline
    Indicator & .630 & .018\\
    Description & \textbf{.698} & \textbf{.186}\\
    Hierarchy & .683 & .154\\
    \hline
    Indicator (MIMIC) & .698 & .041\\
    Description (MIMIC) & .648 & .076\\
    Hierarchy (MIMIC) & .689 & .068\\
    \hline
\end{tabular}
\end{center}
\caption{Code prediction results. AUC and Average precision are computed for the ROC and PR curves respectively in Figures \ref{fig:roc_prc_code_prediction} and \ref{fig:roc_prc_code_prediction_mimic}.}
\label{table:results_code_prediction}
\end{table}

\section{Annotation Statistics}

Here we report general annotation statistics of reference summaries and model validation annotations of past patient reports (Table \ref{table:past_annotation_statistics}) as well as agreement between the targets extracted in Section \ref{section:data_extraction} with those annotated in the future reports of both annotation rounds (Table \ref{table:future_annotation_statistics}).  We also report annotator agreement over 4 instances in Table \ref{table:annotator_agreement}.  In this table, custom columns have no overlap because they were created by the annotators individually.

\begin{table}
\small
\begin{center}
\begin{tabular}{ l c c c c}
    \hline
     & Instances & Patients & Queries & Sentences\\
    \hline
    Reference Summaries & 37 & 35 & 162 & 1840\\
    Model Validation & 12 & 7 & 23 & 900\\
    \hline
\end{tabular}
\end{center}
\caption{Past Report Annotation Statistics.}
\label{table:past_annotation_statistics}
\end{table}

\begin{table}
\small
\begin{center}
\begin{tabular}{ c c c c}
    \hline
     True Positives & False Positives & True Negatives & False Negatives\\
    \hline
    107 & 231 & 160199 & 105\\
    \hline
\end{tabular}
\end{center}
\caption{Target Validation.  We compute overlap between targets produced by the data extraction in Section \ref{section:data_extraction} and those annotated in the future reports by clinicians in both rounds.}
\label{table:future_annotation_statistics}
\end{table}

\begin{table}
\small
\begin{center}
\begin{tabular}{ l c c c}
    \hline
      & Overlapping & Annotator 1 & Annotator 2\\
    \hline
    Query Counts Excluding Custom & 10 & 2 & 18\\
    Custom Query Counts & - & 3 & 3\\
    Sentence Counts (for overlapping queries) & 40 & 13 & 110\\
    \hline
\end{tabular}
\end{center}
\caption{Annotator Agreement for reference summaries on 4 instances spanning 4 patients.  Annotator columns denote counts of items only annotated by that annotator and not the other.}
\label{table:annotator_agreement}
\end{table}

\end{document}